\documentclass[conference, 10pt]{IEEEtran}
\IEEEoverridecommandlockouts 
\usepackage[]{graphicx}
\graphicspath{{Figures/}}
\usepackage{caption}
\usepackage{subfig} 
\usepackage{amsmath}
\usepackage{amssymb}
\usepackage{epsfig}
\usepackage{cite}
\usepackage{color}
\usepackage{balance}
\usepackage{amsmath}
\usepackage{amsfonts} % for the \checkmark command 
\usepackage{graphicx}
\usepackage{pdfpages}
\usepackage[T1]{fontenc} % For less than and greater than sign
\usepackage{array}
\usepackage{url}

\usepackage{color}
\usepackage{wrapfig}
\usepackage{lipsum}
\usepackage[hidelinks]{hyperref}
\usepackage{multirow}
\usepackage{tabularx}
\usepackage{tikz}
\usepackage{nth} % For 5th 6th
\usepackage{algpseudocode} % For algorithm 
\usepackage{algorithm,algpseudocode}
\usepackage{breqn}
\begin{document}
\title{Multi-label Cloud Segmentation\\ Using a Deep Network}
\author{\IEEEauthorblockN{Soumyabrata Dev\IEEEauthorrefmark{3},
Shilpa Manandhar\IEEEauthorrefmark{1},
Yee Hui Lee\IEEEauthorrefmark{1}, and
Stefan Winkler\IEEEauthorrefmark{4}}
\IEEEauthorblockA{\IEEEauthorrefmark{3} ADAPT SFI Research Centre, Trinity College Dublin, Ireland}
\IEEEauthorblockA{\IEEEauthorrefmark{1} School of Electrical and Electronic Engineering, Nanyang Technological University (NTU), Singapore}
\IEEEauthorblockA{\IEEEauthorrefmark{4} School of Computing, National University of Singapore (NUS)}
%\thanks{$^{\dagger}$Authors contributed equally and arranged alphabetically.}
%\thanks{This research is funded by the Defence Science and Technology Agency (DSTA), Singapore.}
%\thanks{The ADAPT Centre for Digital Content Technology is funded under the SFI Research Centres Programme (Grant 13/RC/2106) and is co-funded under the European Regional Development Fund.}
\thanks{Send correspondence to Y.\ H.\ Lee, E-mail: EYHLee@ntu.edu.sg.}
\vspace{-0.6cm}
}

% make the title area
\maketitle

\begin{abstract}
Different empirical models have been developed for cloud detection. There is a growing interest in using the ground-based sky/cloud images for this purpose.  Several methods exist that perform binary segmentation of clouds. In this paper, we propose to use a deep learning architecture (U-Net) to perform multi-label sky/cloud image segmentation. The proposed approach outperforms recent literature by a large margin.

\end{abstract}

\IEEEpeerreviewmaketitle

\section{Introduction}
Clouds play an important role in air-to-ground communication. In particuler, satellite communications via very small aperture terminal operating in the Ka-band (20/30 GHz) can be severely impaired by clouds. Cloud attenuation increases as the operating frequency increases. In the literature, there are several works that discuss efficient methods to detect clouds either using empirical models \cite{Yuan_2016} or via satellite images and ground-based camera images. %These methods use texture descriptors~\cite{XX} and stereo-vision~\cite{XX} based approaches %Başeski and Şenaras used texture descriptors 
%for detecting cloud areas in satellite images. %Wu et al. in ~\cite{XX} used an image matching approach in a stereo vision framework, for the application of terrain extraction. 
However, with the recent advancement ground-based sensing and photogrammetric techniques, ground-based observations using sky cameras are gaining more popularity, owing to the lower operating cost  as well as higher temporal and spatial image resolution. Several methods have been devised and proposed in the literature that attempt to perform a binary detection of labels -- \emph{sky} and \emph{cloud} in ground-based sky camera images. These methods~\cite{Li2011, Souza,ICIP1_2014} typically use a discriminatory color feature in different color spaces and components for the cloud detection.

Recently, Dev et al.\ \cite{dev2015multi} proposed using multi-labels for whole sky images by classifying them further into \emph{sky}, \emph{thin cloud}, and \emph{thick cloud}, by modelling the images as a continuous-valued multi-variate distribution. This approach provided good detection rates for \emph{sky} and \emph{thick cloud}, but \emph{thin cloud} detection rate was poor.

\section{Multi-label Segmentation of Cloud Images}

Currently, with the availability of more powerful computing resources, deep neural networks are used for various tasks remote sensing. Deep convolutional neural nets have shown impressive results for the task of image segmentation. We employ a popular deep learning architecture called U-Net~\cite{ronneberger2015u} for the task of recognizing the different labels in a sky/cloud image.\footnote{The code of all simulations in this paper is available online at \url{https://github.com/Soumyabrata/multilabel-unet}}

 U-Net is a deep convolutional neural network that was originally designed for accurate segmentation of electron microscopy images. We train the U-Net architecture on $128\times128$ resized images from our dataset. The output of the U-Net model is a probabilistic mask, wherein each pixel has a \emph{softargmax} value in the range $[0,1]$. This value indicates the degree of cloudiness of the pixel in the sky/cloud image. Finally, we threshold this probabilistic mask into a ternary map using thresholds of $0.3$ and $0.6$. We use these equally spaced thresholds considering equal misclassification costs to the individual labels.

%\subsection{Proposed Approach}
%{\color{red}Here, we briefly talk about the U-net architecture and how it is used here. } 

\section{Experiments \& Results}
\subsection{Dataset}
%Currently, there exists only a single dataset of sky/cloud images containing multi-label semantic ground-truth maps. This dataset is called HYTA dataset, whose images and corresponding ground-truth labels are generated, in consultation with Chinese Academy of Meteorological Sciences, Beijing. The sky/cloud images of this dataset are captured by two ground-based sky cameras located at Beijing and Conghua, in China. %For the purpose of simplicity, the region around the sun (popularly known as the circumsolar region) is avoided, during the curation of the dataset. 
We use the dataset introduced in our previous work~\cite{dev2015multi}. It is based on the images of the HYTA dataset \cite{Li2011}, from which we generated $32$ cropped sky/cloud images, ensuring that diverse sky scene types are being represented. In consultation with expert from the Meteorological Service Singapore, we also annotated the images with segmentation ground-truth containing multiple labels -- \emph{sky}, \emph{thin cloud} and \emph{thick cloud}.\footnote{More details can be found here: \url{https://github.com/Soumyabrata/HYTA}.} 
%comprising multi-labels of sky/cloud images for our experimental evaluations. 

\subsection{Qualitative Evaluation}

\begin{figure*}[htb]
\centering
\includegraphics[height=0.15\textwidth]{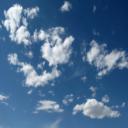}
\includegraphics[height=0.15\textwidth]{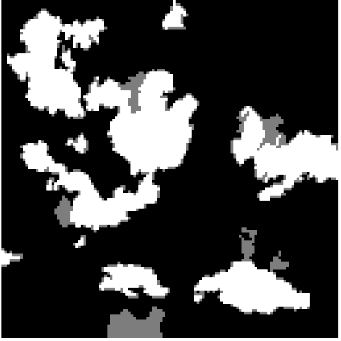}
\includegraphics[height=0.15\textwidth]{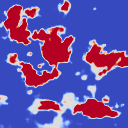}
\includegraphics[height=0.15\textwidth]{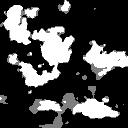}
\includegraphics[height=0.15\textwidth]{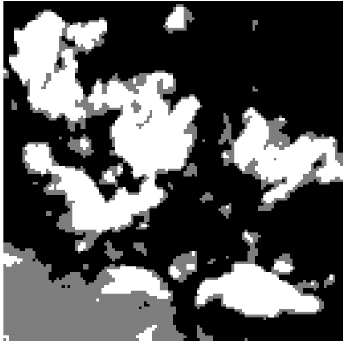}\\
\vspace{1mm}
\includegraphics[height=0.15\textwidth]{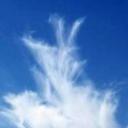}
\includegraphics[height=0.15\textwidth]{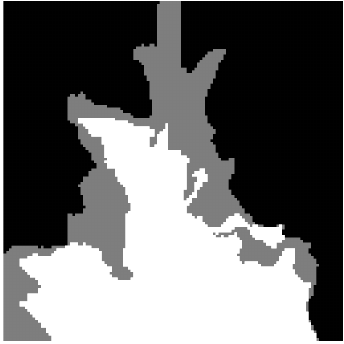}
\includegraphics[height=0.15\textwidth]{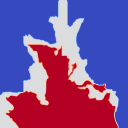}
\includegraphics[height=0.15\textwidth]{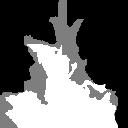}
\includegraphics[height=0.15\textwidth]{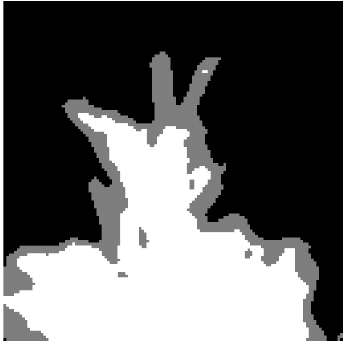}\\
\vspace{1mm}
\includegraphics[height=0.15\textwidth]{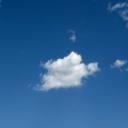}
\includegraphics[height=0.15\textwidth]{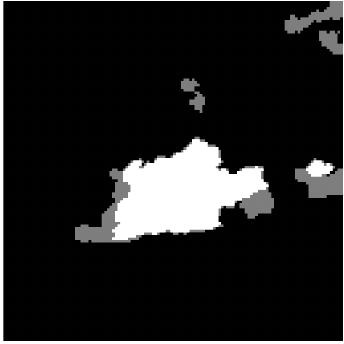}
\includegraphics[height=0.15\textwidth]{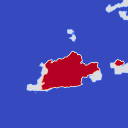}
\includegraphics[height=0.15\textwidth]{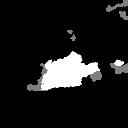}
\includegraphics[height=0.15\textwidth]{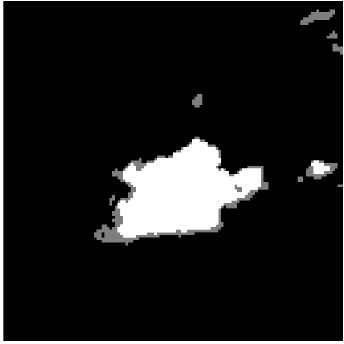}\\
\vspace{-2.5mm}
\subfloat[Input image]{\includegraphics[height=0.15\textwidth]{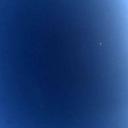}}\ 
\subfloat[Ground truth]{\includegraphics[height=0.15\textwidth]{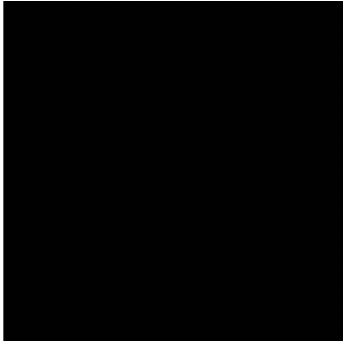}}\ 
\subfloat[Probabilistic U-Net]{\includegraphics[height=0.15\textwidth]{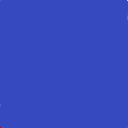}}\ 
\subfloat[Ternary U-Net]{\includegraphics[height=0.15\textwidth]{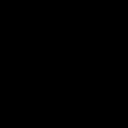}}\ 
\subfloat[Dev et al.\ \cite{dev2015multi}]{\includegraphics[height=0.15\textwidth]{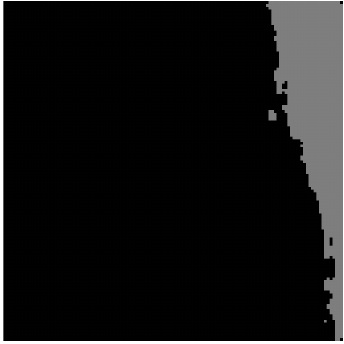}}
\caption{Qualitative evaluation of our proposed approach w.r.t.\ other existing approaches. We use a diverging color map of \emph{coolwarm} for visualizing the probabilistic output of U-Net -- the degree of \emph{redness} indicate higher probability of cloudiness.}
\label{fig:subj-eval}
\vspace{-0.5cm}
\end{figure*}

Figure~\ref{fig:subj-eval} provides a qualitative evaluation of our proposed method using the U-Net architecture. We benchmark our approach against the previous approach  \cite{dev2015multi}. We observe that our approach can efficiently detect the various labels in the sky/cloud image. Most of the improvements are observed in the \emph{thin cloud} regions, which are notoriously difficult to detect using traditional approaches. Also, our deep-learning based approach significantly reduces false positives in the event of a clear sky scene. Furthermore, U-Net also produces a probabilistic output, that can assist the remote sensing analysts to incorporate their subjective judgement in the generated results (although the variation of pixel values within a single cluster is quite small in these examples).

\subsection{Quantitative Evaluation}
In this section, we provide a quantitative evaluation of our proposed approach. We perform a random split of training and testing sets in the ratio of $80:20$. We do not employ any image augmentation techniques during training. The U-Net model is trained on the training set and evaluated on the testing set. We perform this experiment $10$ times, in order to remove any sampling bias. We employ the same experimental procedure to evaluate the Dev et al.\ \cite{dev2015multi} approach. 

We compute the error percentages of the individual labels -- \emph{sky}, \emph{thin cloud}, and \emph{thick cloud} -- across all the testing images. Subsequently, we compute the average error percentage of the different labels over the set of $10$ experiments. Table~\ref{table:compare} summarizes the results. Despite the relatively small size of the dataset, we observe that the performance of our proposed approach is significantly better than the previous work for all the three labels. By far the biggest improvement is observed in the detection of \emph{thin cloud}. 

\begin{table}[t]
\centering
\vspace{-0.05cm}
\small
\caption{Performance evaluation of the proposed approach on the dataset. Values show the misclassification error w.r.t.\ the three labels. We report the error percentage (\%) in the detection of sky-, thin-cloud- and thick-cloud- labels of the testing images.}
\begin{tabular}{crrr}
\hline
Approach & Sky & Thin cloud & Thick cloud \\
\hline 
Dev et al.\ \cite{dev2015multi}  & 15.4\%    & 52.0\%       & 23.4\%  \\
Proposed approach & 7.3\%    & 4.4\%       & 4.4\% \\
\hline
\end{tabular}
\label{table:compare}
\vspace{-0.5cm}
\end{table}

\section{Conclusion}
In this paper, a deep learning architecture is used for multi-label cloud segmentation. Our proposed approach represents a significant improvement over previous work, especially for thin cloud detection. This is particularly interesting for remote sensing applications~\cite{dev2019predicting}, as thin clouds are often mislabelled using conventional approaches.   

%\balance 

\bibliographystyle{IEEEbib}

\end{document}